\documentclass[10pt,twocolumn,letterpaper]{article}

\usepackage{cvpr}              %

\definecolor{cvprblue}{rgb}{0.21,0.49,0.74}
\usepackage[pagebackref,breaklinks,colorlinks,allcolors=cvprblue]{hyperref}

\usepackage{amsmath,graphicx,booktabs,amssymb,multirow}
\usepackage{xcolor, booktabs, multirow, array, placeins, pifont, enumitem, bm, soul, nicefrac, bbold, boldline, tocloft, wrapfig, graphicx, caption, comment, xspace, colortbl}

\newcommand{\xmark}{\ding{55}}
\definecolor{checkmark}{HTML}{40826D}
\definecolor{xmark}{HTML}{E62020}
\newcommand{\ccheck}{{\textcolor{checkmark}{\large\checkmark}}}
\newcommand{\ccross}{{\textcolor{xmark}{\large\xmark}}}

\usepackage[capitalize]{cleveref}
\usepackage[pagebackref,breaklinks,colorlinks]{hyperref}
\def\etal{\emph{et al. }}
\def\eg{\emph{e.g., }}

\newcommand{\method}[1]{{\textbf{PF3Det}}}

\def\paperID{5} %
\def\confName{CVPR}
\def\confYear{2025}

\title{\method~: A Prompted Foundation Feature Assisted Visual LiDAR 3D Detector}

\author{
    Kaidong Li$^{1}$$^\star$\quad Tianxiao Zhang$^{1}$\quad Kuan-Chuan Peng$^{2}$\quad Guanghui Wang$^{3}$
    \vspace{8pt}\\
    $^{1}$University of Kansas\qquad $^{2}$Mitsubishi Electric Research Laboratories\\
    $^{3}$Toronto Metropolitan University
    \\
    {\tt\small {likaidong910@gmail.com} \quad  \tt\small {tianxiao@ku.edu} \quad  \tt\small {kpeng@merl.com} \quad \tt\small {wangcs@torontomu.ca}}
}

\newcommand\blfootnote[1]{%
  \begingroup
  \renewcommand\thefootnote{}\footnote{#1}%
  \addtocounter{footnote}{-1}%
  \endgroup
}

\begin{document}
\maketitle

\blfootnote{$^\star$ This work was done when Kaidong Li was an intern at MERL.}

\begin{abstract}
3D object detection is crucial for autonomous driving, leveraging both LiDAR point clouds for precise depth information and camera images for rich semantic information. Therefore, the multi-modal methods that combine both modalities offer more robust detection results. However, efficiently fusing LiDAR points and images remains challenging due to the domain gaps. In addition, the performance of many models is limited by the amount of high quality labeled data, which is expensive to create. The recent advances in foundation models, which use large-scale pre-training on different modalities, enable better multi-modal fusion. Combining the prompt engineering techniques for efficient training, we propose the Prompted Foundational 3D Detector (\method~), which integrates foundation model encoders and soft prompts to enhance LiDAR-camera feature fusion. \method~ achieves the state-of-the-art results under limited training data, improving NDS by $1.19\%$ and mAP by $2.42\%$ on the nuScenes dataset, demonstrating its efficiency in 3D detection.
\end{abstract}

\section{Introduction}
\label{sec:intro}
3D object detection is an essential task in autonomous driving, involving the localization and classification of objects in 3D space, which is different from 2D object detection \cite{ma2020location}\cite{ma2021miti}\cite{zhang2021six}\cite{zhang2022dynamic} that classifies and localizes the objects in 2D visual data. The LiDAR point clouds provide precise depth information. Early LiDAR-only methods like VoxelNet \cite{zhou2018voxelnet} achieved promising results. However, the sparsity of point clouds makes it difficult to detect small or distant objects. In contrast, high-resolution images from cameras provide rich semantic details, but camera-only data do not have precise distance information, which is why they usually do not match the performance of LiDAR-only approaches. Wang \etal \cite{wang2022probabilistic} find that depth prediction error contributes $20\%$ to the overall error of camera-only models. Therefore, the complementary strength of LiDAR and camera makes multi-modal detection a more robust solution.

\begin{figure}[t]
    \includegraphics[width=1\linewidth]{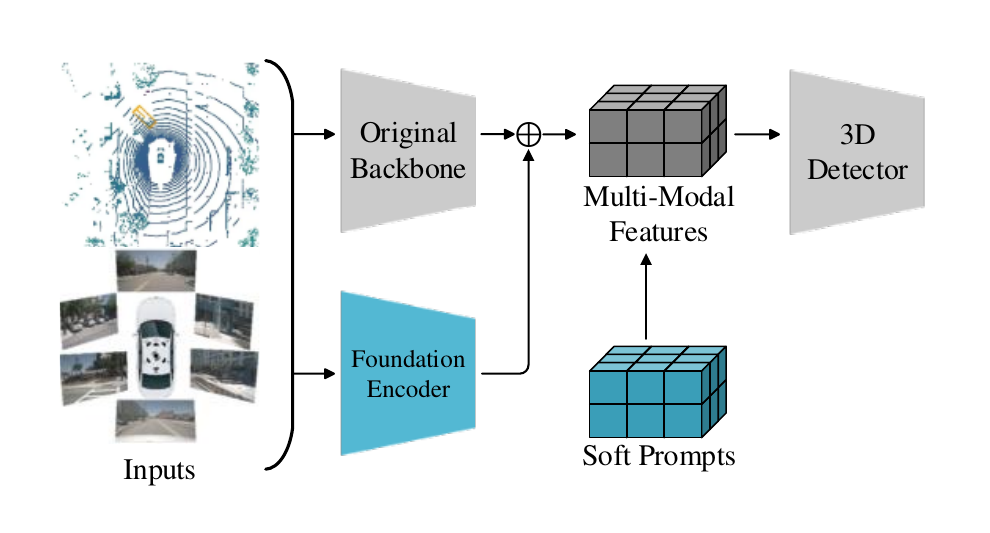}
    \vspace{-7mm}
    \caption{The illustration of the \method~ pipeline. The blue components are the proposed modules.} 
    \label{fig:intro}
\end{figure}

Since the LiDAR-based methods demonstrated stronger performance compared to the camera-based ones, the main stream approach is based on LiDAR detectors, and tries to fuse additional information from the camera modality into the LiDAR models. The main challenge of multi-modal detection is how to fuse the two modalities efficiently. Some multi-modal detectors try to establish point-to-pixel correspondence and query rich camera features from 3D data points \cite{liang2018deep, liang2019multi, bai2022transfusion}. However, the mismatched data densities inevitably result in loss of 2D information. Another paradigm of visual LiDAR detectors first estimates the depth of 2D data and then lifts the 2D pixels into the 3D space \cite{yin2021multimodal, liang2022bevfusion, jiao2023msmdfusion}. These methods have improved the prediction quality thanks to integrating more information from the 2D data. However, lifting a 2D pixel into 3D requires depth estimation, which is still prone to errors \cite{wang2022probabilistic}. Also the feature spaces from the two different modalities are not aligned, thus diminishing the effect of the final feature maps before detection.

Another challenge for existing methods is that they heavily rely on extensive and high-quality annotations \cite{gao2021we, liu2022less, kong2024multi}. Labeling a single scene demands considerable efforts, which makes large-scale labeling impractical for applications such as robotics and autonomous vehicles. Even with existing dataset, Gao \etal \cite{gao2021we} show the performance of the state-of-the-art (SOTA) models are limited by the amount of data. The situation is worse in industry, since many datasets cannot be used for commercial purposes. Therefore, providing a data-efficient 3D detector is crucial.

In recent years, there has been significant progress in developing multi-modal foundation models, which are trained on large-scale datasets and can be adapted for a wide range of downstream tasks \cite{jia2021scaling, radford2021learning, kirillov2023segment, achiam2023gpt, xu2023pointllm}. Their success can be mainly attributed to the capability of large language models (LLMs), which enables massive scaling up on data. The training of LLMs on noisy web-scale data inspires the training on other modalities \cite{radford2021learning}. These models like GPT-3 \cite{brown2020language}, with billions of parameters, have demonstrated impressive performance in zero- and few-shot learning, achieving strong results without requiring task-specific data or parameter updates. The success facilitates the training of vision models to the feature space of LLMs. CLIP uses contrastive pre-training with LLMs to generate a model with promising zero-shot classification and detection performance \cite{radford2021learning}. More modalities soon join the foundation model family, \eg ULIP \cite{xue2023ulip} proposed a method to create language, image, and point triplet data, and released a point encoder that learns to encode a point cloud into a unified feature space with CLIP. 3D-LLM \cite{hong20233d} interacts with 3D scenes using natural language. It is obvious the combined feature encoders can help to fuse features from different modalities more efficiently.

To adapt a pre-trained network to downstream tasks, prompt engineering is a promising research direction. The foundation models have massive amount of parameters, exceeding billions of parameters \cite{liu2023gpt}. It becomes impractical to efficiently fine-tune a foundation model when the data and computational resources are limited \cite{awais2023foundational}. Soft prompts are tensors, which can be fine-tuned during training while keeping other model weights unchanged. It can adapt to different tasks without altering original weights, which helps preserve the foundation models' capabilities. VPT \cite{jia2022visual} introduces a set of vision patches at different layers of a ViT-based model \cite{dosovitskiy2020image}. 

Motivated by the capabilities of foundation models, we proposed Prompted Foundational 3D Detector (\method~), which is a visual LiDAR 3D detector as shown in Fig. \ref{fig:intro}. It takes the encoder from a foundation model to help extract aligned feature maps from different modalities. The multi-modal soft prompts are inserted at multiple levels to assist the fusion of camera and LiDAR features. \method~ achieves the SOTA performance in efficient 3D detection. According to our experiments, \method~ increases nuScenes detection score (NDS) by $1.19\%$ and mAP by $2.42\%$ on the nuScenes dataset \cite{caesar2020nuscenes} under limited training data.

Our contributions can be summarized below:
\begin{itemize}
    \item We propose \method~, a visual LiDAR 3D detection model architecture that can efficiently learn to predict high-quality 3D objects with a small amount of available data.

    \item To achieve efficient learning, our proposed \method~ takes multi-modal foundational features and bridges the modality domain gaps in the bird-eye-view (BEV) stage by incorporating the soft prompts for convolutional layers. 
    
    \item Our proposed \method~ achieves the SOTA performance on the 3D object detection task under limited training data. We run extensive experiments to explore different settings and provide guidance on how to select model parameters.
    
\end{itemize}

\section{Related work}
\subsection{3D visual-LiDAR detection}
3D visual-LiDAR detection \cite{bai2022transfusion, li2022deepfusion, liang2022bevfusion, zhou2023diffusion, xu2023fusionrcnn} has become increasingly popular in autonomous driving and advanced robotics, taking advantages of the complementary strengths of camera and LiDAR sensors. LiDAR provides precise 3D spatial awareness, while cameras provide rich color and texture details. The sensor fusion techniques can be generally divided into three categories, early fusion, intermediate fusion, and late fusion. These categories are based on when the camera and LiDAR features are fused. \textbf{Early fusion} first extracts information from the 2D camera, then adds the information to LiDAR data before LiDAR detectors. Frustum PointNets \cite{qi2018frustum} uses a 2D detector to generate 2D bounding box (bbox) proposals, and then those 2D bboxes are used to produce frustum proposals to help 3D point cloud segmentation. Du \etal \cite{du2018general} proposed similar frustum-based method, but in later stages, implemented a model-fitting algorithm to refine point segmentation using generalized object models. PointPainting \cite{vora2020pointpainting} first runs pixel-level segmentation on 2D images, and passes pixel labels to each points via pixel-to-point correlation. However, early fusion methods are usually sequential, which increases latency \cite{mao20233d}. \textbf{Late fusion} happens after image and point branches generate their 2D and 3D bboxes separately, which means that the image-based and point-based models can run in parallel. Pang \etal proposed CLOCS \cite{pang2020clocs}, which converts 2D and 3D detection candidates into sparse tensors and generates final results from the sparse tensors. However, late fusion does not provide advanced integration of the two modalities, losing the rich semantic features from the other sensor. \textbf{Intermediate fusion} refers to any fusion methods that happen at the stages anywhere between early fusion and late fusion, \eg at backbone stage \cite{xu2018pointfusion, sindagi2019mvx}, at proposal generation stage \cite{chen2017multi, ku2018joint}, \etc. This type of detectors enables deeper integration of two modalities, generates features with higher quality, and witnesses many SOTA 3D detectors \cite{kim20223d, jiao2023msmdfusion, yoo20203d}. But one challenge is that the camera and LiDAR sensor data are heterogeneous, and fusion will even cause noises if it is not done properly. TransFusion \cite{bai2022transfusion} proposed soft-association to better align the data of these two sensors. MSMDFusion \cite{jiao2023msmdfusion} designed two modules, Multi-Depth Unprojection (MDU) and Gated, Modality-Aware Convolution (GMA-Conv), to create alignment and deep fusion between camera and LiDAR's features.

\begin{figure*}[t]
    \includegraphics[width=1\linewidth]{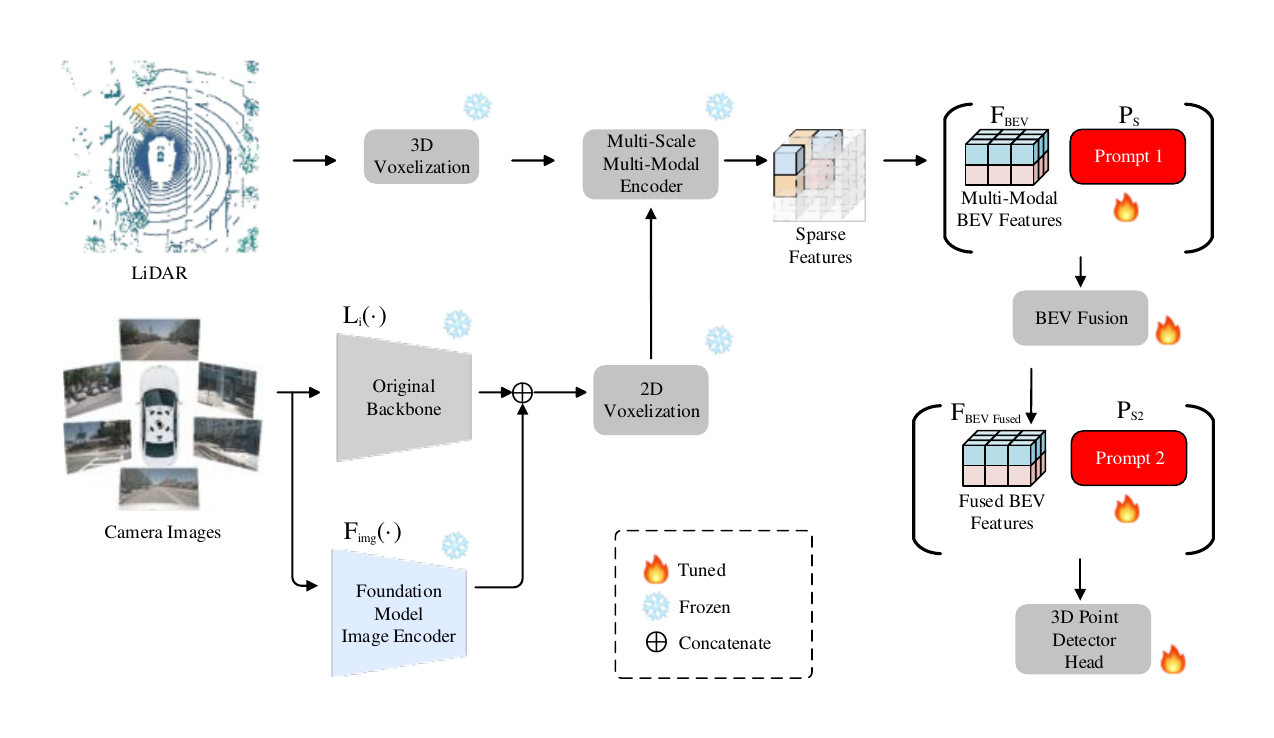}
    \vspace{-10mm}
    \caption{The architecture of our proposed \method~. The Foundational branch is added in parallel with the original image backbone. Multi-modal soft-prompt adapter is inserted at the BEV feature level.} 
    \label{fig:main}
\end{figure*}

\subsection{Multi-modal large language models}
Multi-modal Large Language Models (MLLMs) are designed to achieve a comprehensive understanding across various modalities, including audio \cite{huang2023audiogpt}, images \cite{radford2021learning}, point clouds \cite{xue2023ulip, li2024robust}, \etc. For visual data, Vision Transformers \cite{dosovitskiy2020image}\cite{zhang2025depth}\cite{zhang2024improving} are usually employed as the visual encoders. The key motivation is that integrating the language modality can significantly enhance a model's ability to interpret complex interactions within diverse input modalities \cite{li2019visualbert, xu2023pointllm}. One approach leverages contrastive learning to align representations from different modalities to joint feature spaces \cite{radford2021learning, xue2023ulip}. 
Notable examples, such as CLIP \cite{radford2021learning} and ALIGN \cite{jia2021scaling}, have demonstrated impressive zero-shot generalization abilities, inspiring a range of applications. For instance, PointCLIP \cite{zhang2022pointclip} leverages CLIP for zero-shot 3D multi-view classification. Another approach uses language as a tool for interacting with other modalities, enabling versatile interactions between human and deep learning models \cite{xue2023ulip}. 

To the best of our knowledge, given the capability of MLLMs, there is not yet an MLLM-assisted 3D visual LiDAR detection model even with the existence of the models like Meta Transformer \cite{zhang2023meta}, which aligns a wide range of modalities including 3D point clouds. However, all these MLLMs only align indoor point clouds for simple objects, which are more dense and consistent compared to LiDAR 3D scans. Empirically it is difficult to bridge the domain gap between these 3D point data, which is the main reason such methods do not exist yet.

\subsection{Prompt engineering}
Prompt engineering is a recent promising method to bridge the domain gap between general foundation models and downstream tasks \cite{lester2021power}. This technique gains popularity when large foundation models demonstrate their capability to adapt to a wide range of applications \cite{jia2022visual}. Comparing to regular fine-tuning, where the knowledge of the foundation models is at the risk of catastrophic forgetting, prompt engineering can keep the original foundation models unchanged \cite{xu2023parameter}, and it is proven in many fields to be useful \cite{pritchett2015learning, brown2020language, lester2021power, hu2021lora, jia2022visual, wei2022chain, liu2023gpt}. Generally speaking, prompt engineering can be divided into two categories, hard prompts and soft prompts. \textbf{Hard prompt tuning} involves manually engineered and human interpretable inputs, sometimes in the form of texts \cite{radford2019language, hong20233d}. For example, \textit{instruction prompting} uses explicit instructions to guide the models to generate desired response \cite{radford2019language, hong20233d}. This approach reduces ambiguity and enhances the model's ability to produce accurate and relevant results. \textit{In-context learning} is often used where rapid adaptation is needed \cite{brown2020language, dong2022survey}. This method enables the foundation models to generalize to specific tasks by showing the models just a few examples or prompts, and \textit{chain-of-thought prompting} breaks down complex tasks into a series of simpler steps or instructions within the prompt, allowing the model to process and solve each step sequentially \cite{wei2022chain}. We refer the readers to these survey papers \cite{lester2021power, gu2023systematic} for more details. This category of methods requires careful crafted prompts and is usually labor-intensive \cite{gu2023systematic}. \textbf{Soft prompt tuning} on the other hand uses learnable parameters \cite{hu2021lora, lester2021power, liu2023gpt}. It is a method that has emerged as an effective way to fine-tune large language models (LLMs) for specific tasks by modifying the input prompts rather than adjusting the model's parameters directly. This approach is particularly beneficial when the computational resources or labeled data are limited \cite{gu2023systematic}. These prompt vectors can be added directly to the input or to any model layers \cite{jia2022visual}. During fine-tuning, the pre-trained model will be frozen, and only the prompt vectors are updated to achieve performance improvements.

\section{Prompted foundational 3D detector}
We propose Prompted Foundational 3D Detector (\method~), a prompted foundation-assisted 3D visual LiDAR detector. This method has two modules, a foundational branch, which utilizes the features from a multi-modal foundation model encoder to assist regular image features, and multi-modal soft-prompt adapter, which accepts soft prompts in the visual-LiDAR BEV layers to improve detection performance. The inputs of our detector are LiDAR points and their corresponding single- or multi-view images. Our \method~ outputs 3D detection bounding boxes. The input images first go through a regular detector backbone and a multi-modal foundational image encoder in parallel. Their features are concatenated to produce combined image foundational features. Then these image features and input point clouds are fed into a MSMDFusion-based \cite{jiao2023msmdfusion} fusion architecture to generate multi-modal BEV features. The BEV features are then integrated with the fine-tuned soft prompts to produce the final adapted multi-modal features for the 3D bbox prediction. The overall model architecture is illustrated in Fig. \ref{fig:main}.

\subsection{Foundational branch}
Regular image backbone can be formulated in Eq. \ref{eq:backbone}, where $L_i$ is the $i$-th layer, $W_i$ is the linear projection at residual connection to match dimensions, and $x_i$ is the $i$-th layer output ($x_0$ is the input images). The multi-scale features $x_i$ will be fused with multi-scale point features.

\begin{equation} \label{eq:backbone}
    x_{i} = L_{i}(x_{i-1}) + W_{i} x_{i-1},\ i=1, 2, ..., N 
\end{equation}

Multi-modal foundation models have learnt to extract inputs from different modalities into unified feature spaces. It can extract semantic-rich features to better capture object relations. To help the model extract more high-level information from the input images, we add another branch, the foundational extractor $F(\cdot)$, to obtain the foundational feature vectors, $v=F(x_0),\ v \in \mathbb{R}^{d}$, where $d$ is the dimension of the feature vectors. Then the vectors $v$ is concatenated to the feature maps $x_N$ from the last layer of the regular image backbone. Then the concatenated features are fed into a feature pyramid networks (FPNs) \cite{lin2017feature} to produce the final multi-scale image features. 

\subsection{Multi-modal soft-prompt adapter}
By borrowing the features from the image encoder of a foundation model directly, there will be inevitably some domain gap between its feature vector and the 3D detection features. Therefore, we propose the multi-modal soft-prompt adapter to integrate soft prompts into the detector architecture. 

The original detector's architecture after multi-modal BEV features $\mathbf{F}_{BEV}$ can be formulated as Eq. \ref{eq:arch_orig_end}. $F_{BEV}$ is obtained by converting the point features, virtual point features, and image features together into the BEV space. $F_{BEV}$ is fed into a BEV fusion encoder, $E_{Fusion}$, which is a combination of four different one-layer CNNs in parallel, followed by another CNN layer to fuse the outputs together. The fused features, $\mathbf{F}_{BEV Fused} \in \mathbb{R}^{C \times H \times W}$, then are used to extract multi-scale BEV features, $\mathbf{F}_{MSB feat.}^{i}$ (where $i$ is the scale index), through a BEV backbone $B_{BEV}$ and BEV FPN detector $P_{BEV}$.

\begin{equation} \label{eq:arch_orig_end}
\begin{split}
    &\mathbf{F}_{BEV Fused} = {E}_{Fusion}(\mathbf{F}_{BEV}) \\
    &(\mathbf{F}_{BEV feat.}^i) = {B}_{BEV}(\mathbf{F}_{BEV Fused}) \\
    &(\mathbf{F}_{MSB feat.}^{i}) = {P}_{BEV}(\mathbf{F}_{BEV feat.}^i) \\
    &\mathbf{F}_{BEV}, \quad \mathbf{F}_{BEV Fused} \in \mathbb{R}^{C \times H \times W} 
\end{split}
\end{equation}

\vspace{2mm}
\noindent
\textbf{Single-level prompts.} The prompts are only added to one single layer in the detector's architecture, which requires less computational resources and is faster to train. Only the single-level prompts, $P_{S} \in \mathbb{R}^{C_p^0 \times H \times W}$, are concatenated to the original BEV feature $\mathbf{F}_{BEV}$ to produce a $(C_p^0+C) \times H \times W$ BEV feature map. We modify the input dimension of the BEV encoder ${E}_{Fusion}$ to take the updated BEV feature, denoted as $\hat{E}_{Fusion}$.

The trainable soft prompts have the same spatial dimension as the feature space. The channel size $C_p^0$ is a hyper-parameter depending on the task complexity. During the training process, the weights of $P_{S}$ are relaxed, which will learn the task-specific knowledge. The detailed formulation is shown in Eq. \ref{eq:arch_prompted}, where $\oplus$ is the concatenation operation.

\vspace{2mm}
\noindent
\textbf{Multi-level prompts.} Multi-level prompts are added at different levels of the detector. We proposed to add two levels of soft prompts, at $F_{BEV}$ and $F_{BEV Fused}$. The dimensions of $F_{BEV}$ and $F_{BEV Fused}$ are both $C \times H \times W$. The detailed formulation is summarized in Eq. \ref{eq:arch_prompted}, where $\oplus$ is the concatenation operation.

\begin{equation} \label{eq:arch_prompted}
\begin{split}
    &\mathbf{F}_{BEV Fused} = \hat{E}_{Fusion}(\mathbf{F}_{BEV} \oplus P_S) \\
    &(\mathbf{F}_{BEV feat.}^i) = {B}_{BEV}(Conv(\mathbf{F}_{BEV Fused} \oplus P_{S2}))
\end{split}
\end{equation}

 For the \textit{first level prompts}, we use the same setting as the single level prompts. For the \textit{second level} at $F_{BEV Fused}$, the prompts, $P_{S2}$, are concatenated to the fused BEV features. Then a single layer CNN, $Conv(\cdot)$ is added to align the dimension for original BEV backbone networks. In summary, for the first level soft prompts, we modify the following layers to adapt to the changed channel size, while for second level, a small convolutional layer is added to compress the additional channels.

\section{Experiment}
Using MSMDFusion \cite{jiao2023msmdfusion} as our baseline, we carry out extensive experiments to compare the performance of our proposed \method~ and the baseline. 

\subsection{Dataset}
The nuScenes dataset \cite{caesar2020nuscenes} is a comprehensive, large-scale dataset for advancing autonomous driving research. It provides 1000 driving scenes of high-quality sensor data collected from Boston, Pittsburgh, Las Vegas, and Singapore. These scenes are captured in a variety of conditions, including different times of day, varying weather scenarios, and diverse traffic situations. They are divided into training, validation, and testing sets, with 700, 150, and 150 scenes, respectively. Each scene is 20 seconds, with each frame containing one point cloud and six calibrated images.

The entire dataset has around 390k LiDAR point clouds and 1.4M images, including approximately 1.4M bounding boxes of 23 ground truth categories. The 3D detection performance for nuScenes is measured in mean Average Precision (mAP), and nuScenes detection score (NDS), which is a combination of mAP and other metrics including the quality of box location, size, orientation, \etc. Following MSMDFusion \cite{jiao2023msmdfusion}, we set the voxel size to (0.075m, 0.075m, 0.2m). To evaluate the model performance under limited amount of training data, we randomly sample $5\%$ from the original nuScenes training set, and keep all data in the validation and testing sets.

\subsection{Implementation details}
For the full version of \method~, we use the VoxelNet \cite{zhou2018voxelnet} as the LiDAR backbone. For the image branch, the original image backbone is a ResNet-50 \cite{he2016deep} with FPN \cite{lin2017feature} pre-trained on ImageNet segmentation task \cite{deng2009imagenet} in a Mask-RCNN segmentation network \cite{he2017mask}. We choose the image encoder with ViT-L \cite{dosovitskiy2020image} as the backbone from CLIP \cite{radford2021learning} for the foundational image encoder. The original backbone $L_i(\cdot)$ takes $448\times800$ images as input and produces four feature maps at different scales. The foundation encoder $F(\cdot)$ takes $224\times224$ images and produce a feature vector of size $C_{f feat.}$. The foundational feature vector is expanded to $C_{f feat.} \cdot W \cdot H$ to match the last feature map of the original backbone and concatenated to it. 

For prompts, our final \method~ uses 2-level prompts. Two sets of prompt with size $[100, 180, 180]$ and $[150, 180, 180]$ are concatenated to the BEV features and fused BEV features, respectively.

Following the MSMDFusion \cite{jiao2023msmdfusion} training strategy, at stage 1, we train the LiDAR branch for 20 epochs using the Transfusion \cite{bai2022transfusion}. At stage 2, the foundation model assisted model and the prompted model are trained separately with the LiDAR model for 6 epochs, using a learning rate of $10^{-4}$. At stage 3, the weights from the two multi-modal detectors are integrated, and we continue to train another 6 epochs with a learning rate of $10^{-5}$. At the inference time, Test-Time Augmentation (TTA) or multi-model ensemble are not used for efficiency.

\subsection{Main results}
Table \ref{table:detection_res} shows the results of our main experiments. We first test different settings for each modules individually. The best settings are combined to produce the final \method~.

\begin{table*}
\renewcommand{\arraystretch}{1.2}
\centering
\resizebox{\linewidth}{!}{
\begin{tabular}{cccccccc}
    \toprule
    Exp. ID &Method &Concat-FM &FM Backbone &Prompt Tuning & Prompt Tuning Layers  & NDS & mAP \\
    \midrule
    \centering 1 &MSMDFusion \cite{jiao2023msmdfusion} &\ccross &N/A &\ccross &N/A &  69.23    &  64.50 \\
    \midrule
    \centering 2 &\method~ &\ccheck &ResNet50 \cite{he2016deep} &\ccross &N/A & 68.36 &  62.86 \\
    \centering 3 &\method~ &\ccheck &ViT-L \cite{dosovitskiy2020image} &\ccross &N/A   & 70.21 &  66.15 \\
    \centering 4 &\method~ &\ccross &N/A &\ccheck &1 &  69.90    &  65.66 \\
    \centering 5 &\method~ &\ccross &N/A &\ccheck &2 &  70.00    &  66.69 \\
    \centering 6 &\method~ &\ccheck &ViT-L \cite{dosovitskiy2020image} &\ccheck & 2 &  \textbf{70.42}    &  \textbf{66.92}  \\
    \bottomrule
\end{tabular}
}
\caption{
Overall experimental results on nuScenes validation dataset. MSMDFusion results are reproduced by training on $5\%$ of nuScenes dataset. When we experiment with the foundation model (FM), we use CLIP \cite{wang2022clip} with either ResNet50 or ViT-L as its backbone. Acronyms: Exp.: experiment; Concat-FM: concatenation with FM features.
}
\label{table:detection_res}
\end{table*}

From Table \ref{table:detection_res}. We have three conclusions.
(1) From Exp. ID 1, 3 and 5 results, the proposed two modules, foundation branch and multi-modal soft-prompt adapter, each can improve visual LiDAR detector performance individually. Furthermore, they can work jointly and further enhance the overall performance from Exp. 6. (2) From the experiments of the foundation features (Exp. ID 2 and 3), we can conclude that the output dimension of the foundation encoders has to suit the original detector. Otherwise, it might reduce the detection performance. CLIP \cite{wang2022clip} ResNet50 encoder outputs the feature vectors of length 1024, but CLIP ViT-L produces the features of length 768. The ResNet50 feature vectors have larger feature dimensions, which dilute the original information. (3) The prompts require careful design to balance the parameter increase and NDS/mAP score boost. One layer can achieve the majority of performance gain, $+0.67\%$ NDS. Adding a second layer results in $2.5\times$ the number of prompt weights, but it only gives a $0.1\%$ NDS gain as shown in Exp. ID=5. More detailed analysis of each module and results are shown in the following subsections.

\subsection{Foundation feature assisted branch}
We run foundation feature encoders from different multi-modal foundation models. The results are summarized in Table \ref{table:fab_encoders}.

\begin{table*}
 \renewcommand{\arraystretch}{1.2}
\centering
\resizebox{\linewidth}{!}{
\begin{tabular}{ccccccc}
    \toprule
    Foundation Model &Modality & Foundation Model Backbone & Channel Size & Upsample & NDS & mAP \\
    \midrule
    CLIP \cite{radford2021learning} &image & ViT-L \cite{dosovitskiy2020image}   & 768 & \ccross  & \textbf{70.21} &  \textbf{66.15} \\
    CLIP \cite{radford2021learning} &image & ResNet50 \cite{he2016deep} & 1024 & \ccross & 68.36 &  62.86 \\
    \midrule
    ULIP \cite{xue2023ulip} &point cloud & PointBERT \cite{yu2022point} & 50  & \ccross & 68.42 &  64.33 \\
    ULIP \cite{xue2023ulip} &point cloud & PointBERT \cite{yu2022point} & 100 & \ccross & 67.89 &  63.66 \\
    ULIP \cite{xue2023ulip} &point cloud & PointBERT \cite{yu2022point} & 50  & \ccheck & 69.50 &  65.53 \\
    ULIP \cite{xue2023ulip} &point cloud & PointBERT \cite{yu2022point} & 100 & \ccheck & 69.35 &  65.74 \\
    \bottomrule
\end{tabular}
}
\caption{
Ablation Study with different foundation feature (FM) modality encoders. The column \textit{Channel Size} stands for foundation feature channel size. The column \textit{Upsample} indicates whether upsampling convolutions are used.
}
\label{table:fab_encoders}
\end{table*}

\noindent
\textbf{Foundation image encoder.} As described in the previous section, foundation image features are added to the image branch as shown in Fig. \ref{fig:main}. We tested the two backbones provided by CLIP. The larger feature vector from ResNet50 \cite{he2016deep} dilutes the original image features and decreases the performance.

\begin{figure}[t]
    \includegraphics[width=1\linewidth]{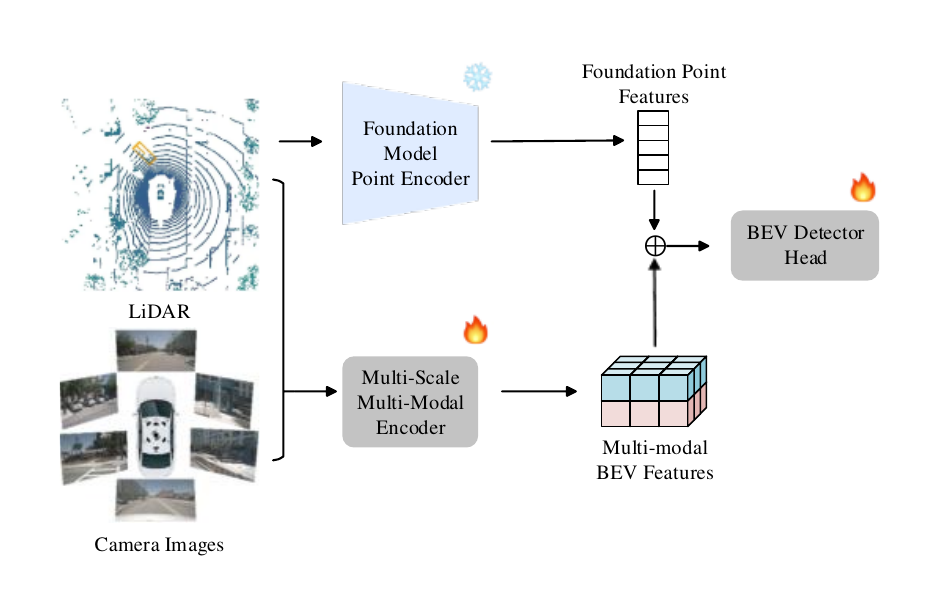}
    \vspace{-7mm}
    \caption{Foundational Point Encoder. The point features are upsampled to match original feature dimensions.} 
    \label{fig:point_encoder}
\end{figure}

\begin{figure*}[t]
    \includegraphics[width=1\linewidth]{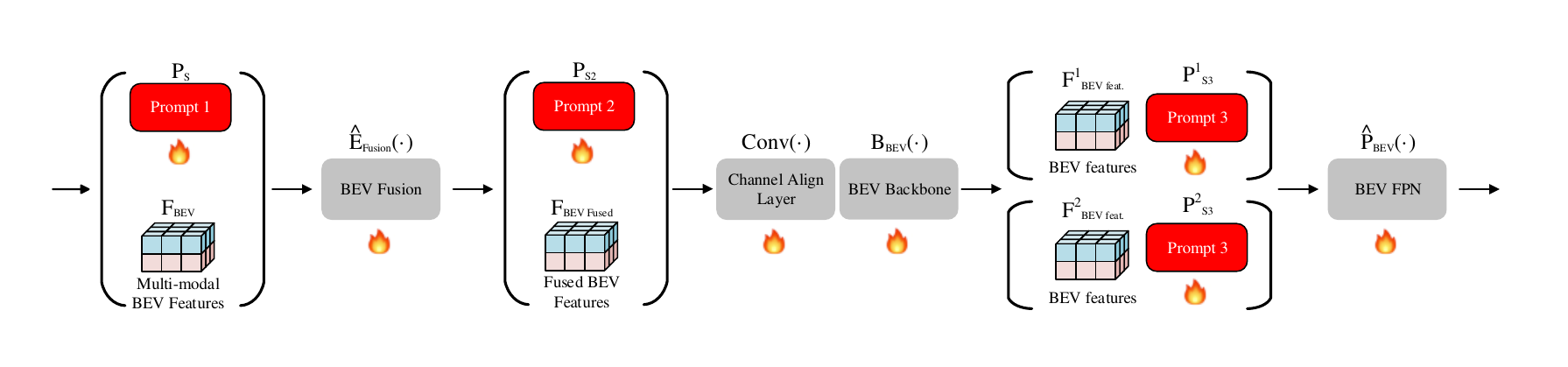}
    \vspace{-8mm}
    \caption{Multi-level multi-modal soft-prompt adapter. Three levels with four sets of soft prompts are tested. And weights after the first prompts are set to be learnable to better fuse features and prompts. } 
    \label{fig:m_prompt}
\end{figure*}

\noindent
\textbf{Foundation point encoder.} We take the point encoder provided by ULIP \cite{xue2023ulip}, which is able to train a model that can encode three modalities into a unified feature space. The point encoder is based on the PointBERT \cite{yu2022point}. Since the features before BEV features in the LiDAR branch are sparse, the foundation point features are integrated to multi-modal BEV features, as shown in Fig. \ref{fig:point_encoder}. To integrate foundation point feature (fp feat.) vectors $[C_{fp feat.}, 1, 1]$ to BEV features $[C_{BEV feat.}, W, H]$, we try to 1) directly repeat vector values to extend to size $[C_{fp feat.}, W, H]$, 2) apply transpose convolution to upsample the vectors. The original BEV features have a channel size of 256, which is the same channel size as the foundation feature vectors. To avoid overwhelming the original information, we add a channel compression layer before concatenation to reduce the foundation point features to size $[C_{comp. fp feat.}, W, H]$. The ablation study is shown in Table \ref{table:fab_encoders}. We have three conclusions: 1) The foundation image features work better compared to the point features. 2) The upsampling layers work better compared to repeating feature vectors. 3) $C_{comp. fp feat.}=50$ has better performance. The reason for conclusion 1 is the domain gap between the clean indoor point clouds and outdoor LiDAR point clouds. The point training data used for ULIP \cite{xue2023ulip} is from ShapeNet \cite{chang2015shapenet}, which is clean and evenly sampled from the object surface compared to the LiDAR points which are highly sparse and uneven.

\subsection{Multi-modal soft-prompt adapter}
We run experiments to search for the optimal prompt structures. The soft prompts are added after the stage of multi-modal BEV features. This allow maximum effect on both the camera and LiDAR modalities. Most of the soft prompts are added to the attention-based networks \cite{jia2022visual}, allowing easy additional prompt patch. However, the BEV feature part of our 3D detector does not have attention-based structures. We design soft prompts specifically for the convolution-based networks. The visual LiDAR 3D detection is a complicated task, and the difficulty to fuse the two modalities is non-trivial. Thus we make all the layers after $F_{BEV}$ learnable and the layers before it to be frozen, as shown in Fig. \ref{fig:main}.

\vspace{2mm}
\noindent
\textbf{Single-level prompts.} First we run experiment to verify the effectiveness of the soft prompts on our foundational encoder assisted networks. Only one layer of soft prompts $P_S$ are added to the feature layer of multi-modal BEV features $F_{BEV}$. The structure is shown in Fig. \ref{fig:m_prompt} without other prompts ($P_{S2}$, $P_{S3}$, $P_{S4}$). 

The multi-modal BEV features $F_{BEV}$ have the dimension of $[256, 180, 180]$, where $256$ is the BEV feature channel size, and $180$ is the BEV width and height. The dimension of the single-level prompt $P_S$ is set to $[C_{P_S}, 180, 180]$ to match the width and height dimension of $F_{BEV}$. Then $P_S$ is concatenated to $F_{BEV}$ to produce a tensor of size $[256 + C_{P_S}, 180, 180]$. The next network layer is a BEV fusion network, as shown in Fig. \ref{fig:m_prompt}. Their input dimensions are adjusted accordingly.

\begin{table}
 \renewcommand{\arraystretch}{1.5}
\centering
\resizebox{\linewidth}{!}{
\begin{tabular}{cr@{\hspace{10mm}}cc}
    \toprule
    Method & \textit{Channel} & NDS & mAP \\
    \midrule
    MSMDFusion \cite{jiao2023msmdfusion}  & - &  69.23    &  64.50 \\
    \midrule
    
    \multirow{7}{*}{\shortstack{\method~\\Single-Level\\Prompts}}    &  10 & 69.22 & 64.63 \\
             &  50 & 69.68 & 65.27 \\
                    & 100 & \textbf{69.90} & \textbf{65.66} \\
                    & 150 & 69.47 & 65.16 \\
                    & 200 & 69.70 & 65.37 \\
                    & 500 & 69.79 & 65.53 \\
                    &1000 & 69.83 & 65.61 \\
    \midrule
    \multirow{4}{*}{\shortstack{\method~\\Multi-Level\\Prompts Reloaded}} & [100, 200, -, -]  & 69.68 & 65.94 \\
    &[100, 150, -, -]&69.76 & 65.89 \\
                & [100, 100, -, -]  & 69.76 & 65.88 \\
                & [100,  50, -, -]  & 69.87 & 66.14 \\
    \midrule
    \multirow{5}{*}{\shortstack{\method~\\Multi-Level\\Prompts}} & [100, 150, -, -]  & \textbf{70.00} & \textbf{66.69} \\
         & [100, 100, -, -]  & 68.97 & 64.50 \\
                & [100,  50, -, -]  & 69.38 & 64.94 \\
                & [150,  75, 38, 75]& 67.66 & 64.20 \\
                & [50,  25, 12, 25] & 67.45 & 64.07 \\
    \bottomrule
\end{tabular}
}
\caption{
\textbf{Detection metrics on different prompt setups.} The column \textit{Channel} indicates the number of channels for each soft prompts. Multi-level prompts reloaded here means the final model is first loaded from a pre-trained single level prompted model and then goes through the training process. Other models in the table are trained from TransFusion \cite{bai2022transfusion} pre-trained weights. The results are reported in $\%$.
}
\label{table:prompt_size}
\end{table}

The results of different prompt channel sizes are shown in Table \ref{table:prompt_size}. The performance of the detector increases as $C_{P_S}$ increases when $C_{P_S} <= 100$. The best performance is achieved when the single prompt channel size $C_{P_S} = 100$. As $C_{P_S}$ increases further, the added prompts start to overwhelm the original BEV features, which has a size of $256$. The detector performance saturates and starts to drop, but we observe that the over-sized prompts do not degrade the performance significantly.

\vspace{2mm}
\noindent
\textbf{Multi-level prompts.} From the prior experiments, we verify that using soft prompts can improve the visual LiDAR 3D detection performance. Further explorations were carried out for multi-level prompts. In total, three levels with four sets of soft prompts, denoted as $P_S$, $P_{S2}$, $P_{S3}^1$, and $P_{S3}^2$, were tested. We show the full architecture in Fig. \ref{fig:m_prompt}.

For $P_S$, we follow the single-level prompt setup. The second level soft prompts, $P_{S2}$, is also set to match the width and height dimension of $F_{BEV Fused}$. Then it is concatenated to the fused BEV features of size $[256, 180, 180]$, to create the combined size of $[256+ C_{P_{S2}}, 180, 180]$. The channel dimension is compressed to the original $256$ by a channel align layer $Conv(\cdot)$, instead of modifying the following BEV backbone layers. The third level of prompts $P_{S3}^i$ are added to the backbone BEV features, $F_{BEV feat.}^i$, after the BEV backbone. As the other prompts, they are concatenated to their corresponding multi-scale BEV features to create features of size $[128+ C_{P_{S3}^1}, 180, 180]$ and $[256+ C_{P_{S3}^2}, 90, 90]$. Like the single level prompts, a modified FPN $\hat{P}_{BEV}(\cdot)$ is used to adapt the changed dimensions. The entire structure shown in Fig. \ref{fig:m_prompt} is set to be trainable.

The experimental results are shown in Table \ref{table:prompt_size}. We run the multi-level prompts with two different settings. In the \textit{multi-level prompts reloaded} section, the models first load the weights from the trained model corresponding to single-level prompts. Only the second level prompts are randomly initialized. The best performance is achieve at $C_{P_{S2}} = 50$. When the channel size of $P_{S2}$ increases to $150$, the performance sees a slight decrease, but remains at the same level. The first significant drop occurs when the channel size increases to $200$, but compared to the single level prompts, they all have slight performance drop. 

Another set of experiments in \textit{multi-level prompts} loads the TransFusion \cite{bai2022transfusion} pre-trained weights as MSMDFusion \cite{jiao2023msmdfusion}. So all prompts are randomly initialized. For two level prompts, the best combination is when $C_{P_S} = 100$ and $C_{P_{S2}} = 150$. Further experiments were carried out to test four levels of prompts. But they all decrease the detection performance significantly by about $1.8\%$. One of the channel sizes, $[50, 25, 12, 25]$, has less prompt weights in total, compared to the best prompt channel sizes $[100, 150, -, -]$, so the four prompts are not overwhelming the original detector. The decrease indicates that the prompts at later layers, \eg multi-scale BEV feature layers, affect the detection performance negatively.

\section{Conclusion}
In this paper, we proposed \method~, a novel foundation feature assisted, prompted visual LiDAR 3D detector. Our proposed modules integrate foundation features learned from multi-modality pre-training and further bridge domain and modality gaps by inserting soft prompts. Extensive experiments support that our \method~ achieves the state-of-the-art results under limited amount of training data.

{
    \small
    \bibliographystyle{ieeenat_fullname}
    \bibliography{main}
}

\end{document}